%% file: ms.tex
\documentclass[conference]{IEEEtran}
\IEEEoverridecommandlockouts
\usepackage{cite}
\usepackage{amsmath,amssymb,amsfonts}
\usepackage{algorithmic}
\usepackage{graphicx}
\usepackage{textcomp}
\usepackage{xcolor}
\usepackage{booktabs}
\usepackage{multirow}
\usepackage{lipsum}
\usepackage{balance}

\def\BibTeX{{\rm B\kern-.05em{\sc i\kern-.025em b}\kern-.08em
    T\kern-.1667em\lower.7ex\hbox{E}\kern-.125emX}}

\input{bibliography/bib.short.def}

\input{src/custom_cmds}

\begin{document}

\title{On-Board Pedestrian Trajectory Prediction Using Behavioral Features}

\author{Phillip Czech$^{1,2}$ \and Markus Braun$^{1}$ \and
    Ulrich Kreßel$^{1}$ \and  Bin Yang$^{2}$
    \thanks{$^{1}$Urban Autonomous Driving Department, Mercedes-Benz AG, Stuttgart, Germany
    {\tt\small phillip.czech@mercedes-benz.com}}%
    \thanks{$^{2}$Institute of Signal Processing and System Theory, University of Stuttgart, Stuttgart, Germany {\tt\small bin.yang@iss.uni-stuttgart.de}}
}

\maketitle

\copyrightnotice

\input{00_abstract}
\input{01_introduction}
\input{02_related_works}
\input{03_method}
\input{04_experiments}
\input{05_conclusion}
\input{06_acknowledgment}

\bibliographystyle{./bibliography/IEEEtran}
\balance
\bibliography{./bibliography/icmla_refs}

\end{document}

%% file: src/custom_cmds.tex
\newcommand\copyrighttext{\footnotesize \textcopyright~2022 IEEE. Personal use of this material is permitted.  Permission from IEEE must be obtained for all other uses, in any current or future media, including reprinting/republishing this material for advertising or promotional purposes, creating new collective works, for resale or redistribution to servers or lists, or reuse of any copyrighted component of this work in other works.
}
\usepackage{tikz}
\newcommand\copyrightnotice{%
    \begin{tikzpicture}[remember picture,overlay]
    \node[anchor=south,xshift=0pt,yshift=14pt] at (current page.south) {\fbox{\parbox{\dimexpr\textwidth-\fboxsep-\fboxrule\relax}{\copyrighttext}}};
    \end{tikzpicture}%
}

%% file: 00_abstract.tex
\begin{abstract}
This paper presents a novel approach to pedestrian trajectory prediction for on-board camera systems, which utilizes behavioral features of pedestrians that can be inferred from visual observations.
Our proposed method, called Behavior-Aware Pedestrian Trajectory Prediction (BA-PTP), processes multiple input modalities, i.e. bounding boxes, body and head orientation of pedestrians as well as their pose, with independent encoding streams.
The encodings of each stream are fused using a modality attention mechanism, resulting in a final embedding that is used to predict future bounding boxes in the image.

In experiments on two datasets for pedestrian behavior prediction, we demonstrate the benefit of using behavioral features for pedestrian trajectory prediction and evaluate the effectiveness of the proposed encoding strategy.
Additionally, we investigate the relevance of different behavioral features on the prediction performance based on an ablation study.
\end{abstract}

%% file: 01_introduction.tex
\section{Introduction}
Anticipating and predicting the future behavior of pedestrians is a key challenge for automated driving.
An intelligent driving system should be able to understand the underlying intentions of pedestrians to be able to safely drive through urban environments \cite{rudenko2020:survey} and prevent accidents.
The prediction of pedestrians’ future behavior brings a lot of challenges, as their behavior is highly variable.
Human drivers use behavioral cues from visual observations to predict the future behavior of pedestrians.
For example, they use the body orientation of pedestrians to assess their moving direction and react accordingly.

In this work, we focus on the problem of pedestrian trajectory prediction from an on-board ego-vehicle perspective, where the goal is to predict the future bounding boxes in the image plane.
This requires capturing relevant motion cues from visual observations in a highly dynamic environment.
Most current approaches for predicting future bounding boxes rely solely on past observed locations \cite{rasouli2019:pie, bhattacharyya2018:bayesian_lstm} of pedestrians to encode their motion history or condition the future positions on their estimated inherent goals \cite{wang2022:sgnet, yao2021:bitrap}.
Considering pedestrians' past observed trajectory is beneficial mainly in cases where they are following a linear movement.
However, pedestrians continuously adjust their paths as they react to the steadily changing environment.
For example, a pedestrian may decide to change his direction when walking over a crossing.
In such cases, more information is needed to predict their future path, especially in the on-board domain where the scene dynamically changes due to the ego-motion of the vehicle.
Observing pedestrians from an ego-centric camera perspective enables capturing their behavioral features that give cues about where they are going to move.
These can be extracted from visual observations and represented in form of features such as the body and head orientation of pedestrians as well as their pose.
Despite that, behavioral features inferred from visual observations are only used in a few works \cite{cao2020:ht_stgcn, sui2021:joint_intention_and_traj_based_on_tranformer} to predict pedestrians' future bounding boxes.

\begin{figure}[t]
\includegraphics[width=1.0\columnwidth]{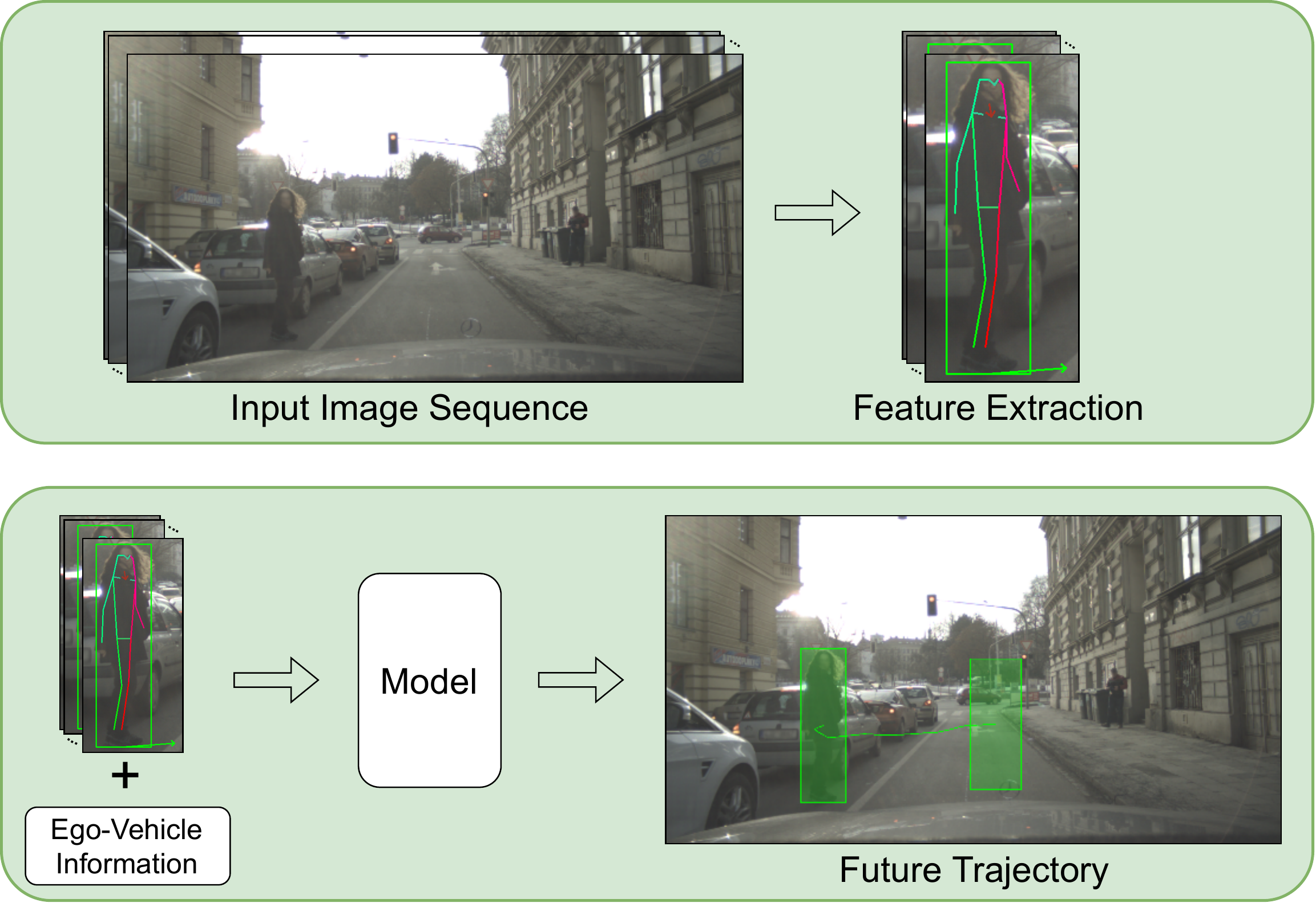}
\caption{Our proposed pedestrian trajectory prediction method (BA-PTP) uses behavioral features such as body and head orientation as well as pose.
These features are combined with ego-vehicle information to predict the future bounding boxes of a pedestrian in the image.}
\label{fig_teaser}
\end{figure}

In this paper, we propose a pedestrian trajectory prediction method for on-board camera systems that uses behavioral features such as body orientation, head orientation as well as pedestrian's pose.
We refer to the method as Behavior-Aware Pedestrian Trajectory Prediction (BA-PTP).
An overview of our proposed method is shown in Fig.~\ref{fig_teaser}.

Our contributions are summarized as follows:
1) We propose BA-PTP, a novel method for pedestrian trajectory prediction in the image plane which utilizes behavioral features of pedestrians.
2) We demonstrate the benefit of including behavioral features in a trajectory prediction model by evaluating BA-PTP on two pedestrian behavior datasets.
3) We show the effectiveness of independently encoding different input modalities to predict more accurate future trajectories.
4) Based on an ablation study we investigate the relevance of different behavioral features on the prediction performance.

%% file: 02_related_works.tex
\section{Related Work}
In recent years, research interest in the domain of pedestrian behavior prediction is growing.
The prediction of pedestrians' future behavior is thereby modeled in two different ways.
First, predicting or classifying the \textit{future action} of a pedestrian, i.e. whether he will cross the street in the near future, often referred to as intention prediction \cite{lorenzo2020:rnn_based_crossing_prediction_using_features, yau2021:graph_sim, rasouli2021:biped, kotseruba2021:pcpa, yao2021:coupling_intent_and_action, sui2021:joint_intention_and_traj_based_on_tranformer, rasouli2017:jaad_baseline, fang2018:pedestrian, cadena2022:pedestrian_graph_plus, liu2020:stip, varytimidis2018:action, yagi2018:future, yang2022:predicting_crossing_with_feat_fusion}.
The second approach predicts the \textit{future trajectory} of a pedestrian for a defined prediction horizon \cite{yao2021:bitrap, cao2020:ht_stgcn, malla2020:titan, rasouli2021:biped, herman2021:bosch_paper, wang2022:sgnet, sui2021:joint_intention_and_traj_based_on_tranformer, bhattacharyya2018:bayesian_lstm, rasouli2019:pie, kooij2019:context_based_path_prediction, ridel2019:understanding, alahi2016:social_lstm, sadeghian2019:sophie, lee2017:desire, salzmann2020:trajectron++, zhao2021:where_are_you_heading, neumann2021:pedestrian_and_ego_veh_traj, yin2021:multimodal_transformer}.
The former approach only gives information about a potential crossing action somewhere in the future, whereas the latter predicts positional information, from which the crossing action can also be derived from the future positions.

Moreover, methods working on the problem of trajectory prediction can be distinguished by the perspective in which they operate.
Pedestrian trajectory prediction in the on-board perspective observes the environment from a moving first-person view.
This results in a dynamically changing scene, because the relative position and size of observed objects depend on the ego-motion of the camera, as the ego-vehicle is also moving.
To address this, some methods try to model the ego-motion by incorporating ego-vehicle information \cite{cao2020:ht_stgcn, malla2020:titan, rasouli2021:biped, bhattacharyya2018:bayesian_lstm, rasouli2019:pie, neumann2021:pedestrian_and_ego_veh_traj, yin2021:multimodal_transformer}.

When predicting the trajectory from a bird's eye perspective, the environment is observed from a top-down view, where the positions of objects are represented by global coordinates and are not dependent on the ego-motion of the camera.
This perspective allows to better model interactions \cite{alahi2016:social_lstm, sadeghian2019:sophie, lee2017:desire, salzmann2020:trajectron++}
as relative distances between objects can be inferred from the global coordinates.
This is rarely done in the camera-based on-board domain \cite{liu2020:stip, malla2020:titan, rasouli2021:biped} where reliable depth information is missing.

Nowadays, recurrent architectures \cite{rasouli2019:pie, bhattacharyya2018:bayesian_lstm, kotseruba2021:pcpa} as well as CVAEs \cite{wang2022:sgnet, herman2021:bosch_paper, yao2021:bitrap} are prevalent in state-of-the-art pedestrian behavior prediction and are used in the majority of recent works.
Other approaches model spatio-temporal features by first extracting features using CNNs \cite{lorenzo2020:rnn_based_crossing_prediction_using_features, malla2020:titan, kotseruba2021:pcpa} or graph structures \cite{liu2020:stip, cao2020:ht_stgcn, yau2021:graph_sim} and processing them later on with RNNs.
Recently, transformers \cite{yin2021:multimodal_transformer, sui2021:joint_intention_and_traj_based_on_tranformer}, as well as goal-driven \cite{wang2022:sgnet, yao2021:bitrap, zhao2021:where_are_you_heading} and attention-based \cite{rasouli2019:pie, kotseruba2021:pcpa, yang2022:predicting_crossing_with_feat_fusion} approaches, have also gained growing interest.

Related work may also be differentiated by the information used as input to predict the future behavior of pedestrians.
The pedestrian's motion history is the most fundamental information as almost all methods rely on the past motion of the pedestrian independent of the domain
\cite{yao2021:bitrap, lorenzo2020:rnn_based_crossing_prediction_using_features, cao2020:ht_stgcn, malla2020:titan, yau2021:graph_sim, rasouli2021:biped, kotseruba2021:pcpa, herman2021:bosch_paper, wang2022:sgnet, yao2021:coupling_intent_and_action, sui2021:joint_intention_and_traj_based_on_tranformer, bhattacharyya2018:bayesian_lstm, rasouli2019:pie, kooij2019:context_based_path_prediction, ridel2019:understanding, yagi2018:future, alahi2016:social_lstm, sadeghian2019:sophie, lee2017:desire, salzmann2020:trajectron++, zhao2021:where_are_you_heading, neumann2021:pedestrian_and_ego_veh_traj, yang2022:predicting_crossing_with_feat_fusion},
This information includes bounding boxes for on-board methods or global coordinates for methods in bird's eye view. 
Additionally, features like 
the distance to the ego-vehicle \cite{herman2021:bosch_paper, kooij2019:context_based_path_prediction},
semantic and contextual information \cite{yau2021:graph_sim, rasouli2021:biped, herman2021:bosch_paper, sui2021:joint_intention_and_traj_based_on_tranformer, liu2020:stip, kooij2019:context_based_path_prediction, lee2017:desire, salzmann2020:trajectron++, yang2022:predicting_crossing_with_feat_fusion},
as well as visual or appearance features \cite{lorenzo2020:rnn_based_crossing_prediction_using_features, cao2020:ht_stgcn, malla2020:titan, kotseruba2021:pcpa, yao2021:coupling_intent_and_action, rasouli2017:jaad_baseline, cadena2022:pedestrian_graph_plus, bhattacharyya2018:bayesian_lstm, rasouli2019:pie, liu2020:stip, neumann2021:pedestrian_and_ego_veh_traj, yang2022:predicting_crossing_with_feat_fusion}
are used as inputs to encode the past motion and behavior of pedestrians.

Behavioral cues such as
body or head orientation \cite{yau2021:graph_sim, herman2021:bosch_paper, sui2021:joint_intention_and_traj_based_on_tranformer, ridel2019:understanding},
awareness \cite{rasouli2017:jaad_baseline, kooij2019:context_based_path_prediction, varytimidis2018:action},
pose of the pedestrian \cite{cao2020:ht_stgcn, kotseruba2021:pcpa, fang2018:pedestrian, cadena2022:pedestrian_graph_plus, yagi2018:future, yang2022:predicting_crossing_with_feat_fusion}
and pedestrians' intention as the wish to cross the street \cite{cao2020:ht_stgcn, rasouli2019:pie, yao2021:coupling_intent_and_action}
are also used to better understand and model pedestrians' behavior.

However, in the domain of on-board pedestrian trajectory prediction in the image plane, such behavioral features are rarely used.
\cite{cao2020:ht_stgcn} uses pedestians' poses to improve the prediction of their intention module, resulting in only a marginal benefit for the trajectory prediction on PIE \cite{rasouli2019:pie}.
The body orientation of pedestrians is used in \cite{sui2021:joint_intention_and_traj_based_on_tranformer} to simultaneously predict intention and trajectory, but only benefits intention prediction, whereas this reduces trajectory prediction performance.

In contrast to the aforementioned methods, we show how to incorporate behavioral features such as the body and head orientation of pedestrians as well as their pose to benefit the prediction of future trajectories in the image plane.
We use independent encoding streams for each input modality and fuse the learned embeddings with an attention mechanism to better encode the motion history of pedestrians and predict more accurate future trajectories.

%% file: 03_method.tex
\section{Method}
\begin{figure*}[tb]
\includegraphics[width=1.0\textwidth]{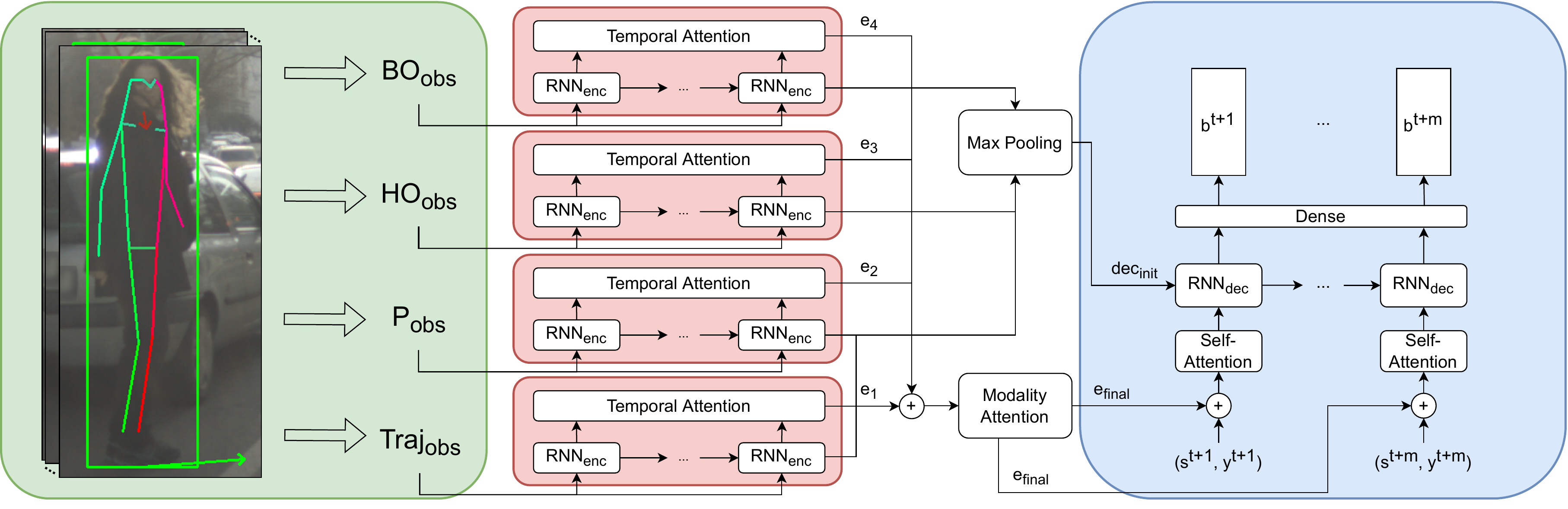}
\caption{
Diagram of our proposed method BA-PTP.
The inputs to the method are bounding boxes and behavioral features such as pedestrian pose, as well as head and body orientation for a defined observation horizon.
Each input modality is processed by an independent encoding stream and the outputs of each stream are fused using a modality attention module, resulting in a final embedding vector.
For each prediction timestep, we concatenate the final embedding vector with vehicle odometry information and pass this through a self-attention unit.
This is then used as the decoder inputs to predict future bounding boxes in the image.
}
\label{fig_method}
\end{figure*}

\subsection{Problem Formulation}
We formulate the problem of forecasting the behavior of pedestrians as the prediction of future trajectories in terms of image positions based on observations of real-world traffic scenes from the on-board ego-vehicle perspective.
Pedestrians' trajectories are thus represented as a sequence of bounding boxes defining the top-left and bottom-right pixel coordinates in the image: $b=\{x_{tl}, y_{tl}, x_{br}, y_{br}\}$.

At a certain timestep $t$ we aim to predict the future trajectory $Traj_{fut}=\{b_{i}^{t+1}, \dots, b_{i}^{t+m}\}$ of a pedestrian $i$ for a prediction horizon $m$, given the pedestrian's observed past trajectory $Traj_{obs}=\{b_{i}^{t-n+1}, \dots, b_{i}^{t}\}$ for an observation horizon of length $n$ as well as observed behavioral features of the pedestrian:
\begin{itemize}
    \item body orientation $BO_{obs}=\{bo_{i}^{t-n+1}, \dots, bo_{i}^{t}\}$,
    \item head orientation $HO_{obs}=\{ho_{i}^{t-n+1}, \dots, ho_{i}^{t}\}$,
    \item pose $P_{obs}=\{p_{i}^{t-n+1}, \dots, p_{i}^{t}\}$, where $p$ is a 34D vector defining the pixel coordinates of a 17-joint skeleton.
\end{itemize}
Additionally, we make use of the future ego-vehicle information, i.e. speed $S_{fut}=\{s_{i}^{t+1}, \dots, s_{i}^{t+m}\}$ as well as the yaw rate $Y_{fut}=\{y_{i}^{t+1}, \dots, y_{i}^{t+m}\}$.

\subsection{Architecture}
Pedestrians moving along a street provide many hints about where they are going to move, which can be inferred from visual observation.
These hints can be represented as behavioral features such as pedestrians' body and head orientation as well as their pose.
To account for this, we propose a method that uses this information as explicit features in addition to their motion history.

As depicted in Fig.~\ref{fig_method}, our method consists of a Recurrent Neural Network (RNN) encoder-decoder architecture, where we process each input modality through an independent encoder, namely a \textit{bounding box stream}, a \textit{pose stream}, a \textit{head orientation stream} and a \textit{body orientation stream}.
This encoding scheme is inspired by \cite{kotseruba2021:pcpa}.
We apply the same structure for each encoding stream $e$ as follows:
The RNN encoder processes the input sequence and produces hidden states for each observed timestep.
We then apply the temporal attention mechanism from \cite{kotseruba2021:pcpa} on the hidden states.
The role of the attention mechanism is to give higher importance to particular timesteps in the observation horizon compared to other ones, focusing on the most relevant information.
The resulting embedding vectors $e_e$ of each encoding stream are concatenated and the modality attention module from \cite{kotseruba2021:pcpa} produces the final embedding vector $e_{final}$ by assigning importance to the different modality inputs.

The final embedding created by our encoding module contains a latent representation of the motion history of the pedestrian, which can be used to predict the future trajectory.
To get a joint representation as input to the trajectory predictor, we concatenate the embedding vector $e_{final}$ with odometry information $(s^{t+j}, y^{t+j})$, i.e. speed and yaw rate, for every future timestep ${t+j}$.
We pass this through a self-attention unit from \cite{rasouli2019:pie} to target the features of the encoding that are most relevant for a certain prediction timestep.
We then use this as the decoder inputs, resulting in hidden states for every future timestep in the prediction horizon.
To generate the final predictions, we pass the decoder outputs through a dense layer that regresses the hidden states for every future timestep into bounding box predictions $Traj_{fut}=\{b_{i}^{t+1}, \dots, b_{i}^{t+m}\}$.

We calculate the decoder's initial state vector $dec_{init}$ by taking the elementwise maximum of the final hidden states $h_e^t$ of all encoding streams, with $e$ being the number of an encoding stream:
\begin{equation}
dec_{init} = \max_{1 \leq j \leq d} h_{ej}^t,
\label{eq_e_final}
\end{equation}
where $d$ is the dimensionality of the hidden states.
We found this to result in the best performance.

\subsection{Loss Function}
Our method is trained in a supervised manner.
As the loss function, we use the root mean squared error (RMSE) between the predicted bounding boxes $\hat{b}$ and the ground truth $b$ for each of the $I$ training samples with prediction horizon $m$:
\begin{equation}
RMSE = \sqrt{\frac{1}{I} \sum_{i=1}^{I} \sum_{j=1}^{m} \|b_i^{t+j} - \hat{b}_i^{t+j}\|^2}
\label{eq_loss}
\end{equation}

%% file: 04_experiments.tex
\section{Experiments}

\subsection{Datasets}
We perform experiments on two different datasets:
\paragraph{PIE \cite{rasouli2019:pie}}
PIE contains on-board camera videos recorded at 30 fps with a resolution of 1920 x 1080 px during daytime in Toronto, Canada.
Annotations include pedestrian bounding boxes with track ids and occlusion flags, as well as vehicle information such as speed and yaw from an on-board diagnostics sensor.
The dataset contains $1\,842$ pedestrians, divided into train, validation and test sets by ratios of $50\%$, $10\%$ and $40\%$ respectively.

\paragraph{ECP-Intention}
The ECP-Intention dataset was created by selecting recordings collected for the EuroCity Persons (ECP) dataset \cite{braun2019:ecp}. 
The data was recorded with a two-megapixel camera (1920 x 1024 px) mounted behind the windshield at 20 fps in 31 different cities in Europe.
227 sequences from 27 cities with an average length of 16 seconds were selected.
Each sequence was manually labeled at 5 fps (every 4th image) providing pedestrian bounding boxes with track ids and occlusion/truncation flags.
Additionally, for all pedestrians, their body and head orientation relative to the line of sight of the camera in the range $[0, 360)$ was annotated.
Thus, a value of $0$ degrees corresponds to an orientation directly towards the camera.
For each image, odometry information is provided captured by the vehicle's sensors and an additional IMU/GNSS sensor: the vehicle's speed and yaw rate.

Overall, ECP-Intention consists of $17\,299$ manually labeled images with $133\,030$ pedestrian bounding boxes containing $6\,344$ unique pedestrian trajectories. 
The data is split into train, validation and test subsets by ratios of $55\%$, $10\%$ and $35\%$ respectively.
The dataset is not published yet.

Since there are images at 20 fps available for ECP-Intention, we extend the 5 fps manual annotations to 20 fps by linear interpolation between two hand-labeled frames, called \textit{keyframes}.
This increases the amount of data in the observation horizon and enables us to better train our method since it can rely on more information for encoding the motion history.
In addition to that, we can generate more data samples.

\subsection{Implementation}
\label{sec_implementation}
We use Gated Recurrent Units (GRUs) with 256 hidden units and \textit{tanh} activation for the encoding streams as well as for the decoder.
Observation ($n$) and prediction horizon ($m$) are set based on the dataset used:
For PIE we follow \cite{rasouli2019:pie} and use $0.5$ seconds of observations ($n=15$) and predict $1.5$ seconds ($m=45$).
We sample tracks with a sliding window approach and a stride of 7.
For ECP-Intention we observe $0.6$ seconds ($n=13$) and predict $1.6$ seconds ($m=32$).
Tracks are sampled with a stride of 2.
If there are missing data frames within a sequence of the same pedestrian (e.g. due to occlusion), we split the sequence into multiple tracks to ensure consecutive tracks during training and testing.
We train every model for $80$ epochs using the Adam \cite{kingma2014:adam} optimizer with a batch size of $128$ and $L_2$ regularization of $10^{-4}$.
After each temporal attention block, we use a dropout of $0.5$ following \cite{kotseruba2021:pcpa}.
The initial learning rate for ECP-Intention and PIE is set to $5 \times 10^{-4}$ and $10^{-3}$ respectively.
If the validation loss has not improved for 5 epochs during training, the learning rate is reduced by a factor of 5.

Since we do not have annotated poses for ECP-Intention and no annotations for behavioral features at all for PIE, we need to generate the data beforehand.
To infer the pose and body orientation of pedestrians we use the following architecture:
For detection and pose estimation of pedestrians, we use VRU-Pose SSD \cite{kumar2021:pose_ssd} extended by the orientation loss used in Pose-RCNN \cite{braun2016:pose_rcnn}.
This way we simultaneously predict the pose and body orientation of pedestrians.
The model is trained on a combination of three datasets: ECPDP \cite{braun2021:simple_pair_pose}, TDUP \cite{wang2021:urbanpose} as well as additional non-public data.
To associate these detections with the ground truth boxes of ECP-Intention and PIE, we match every ground truth bounding box with the detection with the highest \textit{Intersection over Union (IoU)}.
Missing attribute values are ignored by our method using a masking layer.

\subsection{Data Preparation}
For the trajectory input $Traj_{obs}$ we use the ground truth bounding boxes.
The reason for that is to make the trajectory prediction independent of a pedestrian detector's performance as in the majority of previous works \cite{yao2021:bitrap, rasouli2019:pie, rasouli2021:biped, malla2020:titan, wang2022:sgnet}.
We normalize $Traj_{obs}$ by subtracting the first bounding box from the whole track, i.e. converting the absolute to relative pixel coordinates, as it is done in \cite{rasouli2019:pie}.

The pedestrian's pose $p_i$ is normalized by dividing each x-coordinate by the image width and each y-coordinate by the image height.
%
The values for body and head orientation are normalized to the range $[0, 1)$.
For odometry information, we do not apply any pre-processing.

\subsection{Metrics}
We use the standard metrics \cite{rasouli2019:pie, wang2022:sgnet, bhattacharyya2018:bayesian_lstm} for evaluation on both datasets PIE and ECP-Intention:
the mean squared error (\textbf{MSE}) of the bounding boxes' upper-left and lower-right corner is calculated over the different prediction horizons as well as the MSE of the bounding box centers over the whole prediction horizon (\textbf{C\textsubscript{MSE}}) and only for the final timestep (\textbf{CF\textsubscript{MSE}}).
For ECP-Intention we calculate the metrics only for the keyframes to evaluate the performance only on the hand-labeled data.
To this end, we restrict test samples to end with keyframes during data generation enabling us to measure prediction performance exactly $1.6$ seconds into the future.

\subsection{Quantitative Results}
\label{sec_quantitative_results}
We average the results of all models over four different experiment runs.
In the following sections, we report the average performance with standard deviation in integer pixel errors by mathematical rounding.
\paragraph{Comparison with SOTA}
We perform experiments on the ECP-Intention dataset comparing our method to the state-of-the-art for on-board pedestrian trajectory prediction in the image plane.
We compare to PIE\textsubscript{traj} \cite{rasouli2019:pie} and it's extension PIE\textsubscript{traj+speed} \cite{rasouli2019:pie}, which additionally uses ground truth ego-vehicle speed information.
We also compare to SGNet \cite{wang2022:sgnet} which is a goal-driven approach and currently the best performing deterministic model on the PIE dataset.

In Table~\ref{table_ecp_results} we see that our method outperforms all proposed baselines on all metrics by a considerable margin even without taking any behavioral features into account (BA-PTP\textsubscript{BB}).
This can be contributed to the used yaw rate of the ego-vehicle improving prediction performance in turning scenarios, which the compared baselines do not use.
This result implies that the incorporation of the ego-vehicle's yaw rate is a prerequisite to deal with the ego-motion of the camera in the sequences of ECP-Intention.
This assumption can be confirmed by comparing the performance to our model that does not use the yaw rate (BA-PTP\textsubscript{BB-Y}).
When we add the yaw rate to our model the performance increases by 39\% for MSE-0.8s and for longer predictions by up to 57\% for MSE-1.6s.
We use future ground-truth odometry in this work as this information is assumed to be known for the prediction horizon (e.g. based on an automated vehicle's planner).
%
%
The impact of predicted odometry information from an ego-motion model on the accuracy of trajectory prediction is left for future work.

The best performing model uses body orientation and pose of pedestrians in addition to bounding boxes and odometry information for predicting future trajectories (BA-PTP\textsubscript{BB+BO+P}).
The trajectory prediction performance increases by 22\% in MSE-1.6s, 23\% in C\textsubscript{MSE} and 27\% in CF\textsubscript{MSE} compared to only using bounding boxes and odometry information.
This shows the benefit of using behavioral features of pedestrians for pedestrian trajectory prediction.

\begin{table}[tb]
\centering
\caption{Trajectory prediction results on the ECP-Intention dataset
 \\ (BB=Bounding Boxes, BO=Body Orientation,
 \\ HO=Head Orientation, P=Pose)
 \\ * denotes no yaw rate used
}\label{table_ecp_results}
\begin{tabular}{@{}lccccc@{}}
\toprule
\multirow{2}{*}{Model} & \multicolumn{2}{c}{\begin{tabular}[c]{@{}c@{}}MSE\\ Avg (Std)\end{tabular}} & \begin{tabular}[c]{@{}c@{}}C\textsubscript{MSE}\\ Avg (Std)\end{tabular} & \begin{tabular}[c]{@{}c@{}}CF\textsubscript{MSE}\\ Avg (Std)\end{tabular} \\
                      & 0.8s                                    & 1.6s                  & 1.6s                                   & 1.6s                          \\ \midrule
PIE\textsubscript{traj} \cite{rasouli2019:pie}* & 434$\pm$24    & 2216$\pm$71           & 2088$\pm$70                            & 6785$\pm$136                  \\
PIE\textsubscript{traj+speed} \cite{rasouli2019:pie}* & 417$\pm$10 & 2120$\pm$63        & 1997$\pm$67                            & 6516$\pm$235                  \\
SGNet \cite{wang2022:sgnet}* & 390$\pm$8                        & 2137$\pm$94           & 2033$\pm$92                            & 6708$\pm$330                  \\ \midrule
BA-PTP\textsubscript{BB-Y}* & 435$\pm$21                        & 2339$\pm$63           & 2202$\pm$54                            & 7290$\pm$154                  \\
BA-PTP\textsubscript{BB}          & 261$\pm$3                   & 987$\pm$15            & 885$\pm$12                             & 2699$\pm$90                   \\ \midrule
BA-PTP\textsubscript{BB+HO}       & 249$\pm$1                   & 882$\pm$7             & 783$\pm$9                              & 2355$\pm$20                   \\
BA-PTP\textsubscript{BB+BO}       & 244$\pm$3                   & 876$\pm$13            & 780$\pm$16                             & 2356$\pm$62                   \\
BA-PTP\textsubscript{BB+P}        & 240$\pm$9                   & 845$\pm$26            & 746$\pm$23                             & 2250$\pm$78                   \\ \midrule
BA-PTP\textsubscript{BB+BO+HO}    & 247$\pm$2                   & 866$\pm$10            & 768$\pm$11                             & 2303$\pm$42                   \\ 
BA-PTP\textsubscript{BB+HO+P}     & 241$\pm$6                   & 790$\pm$19            & 697$\pm$21                             & 2024$\pm$65                   \\ 
\textbf{BA-PTP\textsubscript{BB+BO+P}} & \textbf{233$\pm$6}     & \textbf{768$\pm$32}   & \textbf{680$\pm$35}                    & \textbf{1966$\pm$91}          \\
BA-PTP\textsubscript{BB+BO+HO+P}  & 236$\pm$2                   & 777$\pm$13            & 685$\pm$14                             & 2003$\pm$46                   \\ \midrule
\textit{BA-PTP\textsubscript{BB +infer. BO}} & 247$\pm$2        & 890$\pm$11            & 790$\pm$13                             & 2391$\pm$48                   \\
\textit{BA-PTP\textsubscript{BB +infer. BO+P}} & 233$\pm$2      & 787$\pm$19            & 693$\pm$20                             & 2029$\pm$66                   \\ \bottomrule
\end{tabular}
\end{table}

\paragraph{Relevance of different behavioral features}
We perform an ablation study on ECP-Intention to investigate which behavioral features contribute most to the prediction performance.
The results of our method when we use only a subset of the behavioral features are also shown in Table~\ref{table_ecp_results}.
Rows 6-8 show how the performance changes when we use only one of the features (\textit{BO, HO, P}) in addition to bounding box and odometry data.
Each of the behavioral features improves the trajectory prediction.
Pose has the highest influence on the prediction performance reducing the MSE relative to only using bounding boxes by 8\% for 0.8s prediction and 14\% for 1.6s prediction.

Rows 9-12 show the results of our method when combining multiple behavioral features.
Based on this ablation study, we can draw multiple conclusions.
Combining two behavioral features results in better performance than solely using one additional feature indicating that the features complement each other.
The combination of body orientation and pose results in the best average prediction performance on the ECP-Intention dataset.
However, if we also add the head orientation, i.e. use all behavioral features, the prediction performance decreases slightly.
We lose 1\% across all metrics.
This indicates, that the information about the head orientation might already be captured in the combination of pose and body orientation features.

Additionally, the \textit{italic rows} in Table~\ref{table_ecp_results} show experiments, where we use inferred body orientation instead of the ground truth annotations as input for both training and inference.
We can see that the performance decreases by a few points across all metrics compared to the ground truth counterparts which can be explained by the introduced noise from using inferred values.
Still, we see significant benefits compared to only using bounding boxes and odometry information.

\paragraph{Effect of independent encoding streams}
In Table~\ref{table_encodings} we compare our attention-based modality fusion (\textit{Independent}) with a model variant, where we concatenate all input features first and process them with a single encoding stream (\textit{Concat}).
Thus, this model variant solely uses temporal attention and no modality attention.
The only additional change we made is that we initialize the decoder's initial state with zeros, which we found to result in the best performance for the \textit{Concat} encoding strategy.
We show this ablation study for two models: BA-PTP\textsubscript{BB+BO+P} and BA-PTP\textsubscript{BB+BO+HO+P}.
Our \textit{Independent} encoding strategy achieves better results for both models in comparison with the \textit{Concat} encoding strategy.
For example, for BA-PTP\textsubscript{BB+BO+P} MSE-1.6s, C\textsubscript{MSE} and CF\textsubscript{MSE} improve by 3\%, 1\% and 2\%, respectively.

\begin{table}[tb]
\caption{Comparison of different encoding strategies
\\ (C=Concat, I=Independent)}\label{table_encodings}
\begin{center}
\begin{tabular}{@{}lccccc@{}}
\toprule
\multirow{2}{*}{Model} & \multicolumn{2}{c}{\begin{tabular}[c]{@{}c@{}}MSE\\ Avg (Std)\end{tabular}} & \begin{tabular}[c]{@{}c@{}}C\textsubscript{MSE}\\ Avg (Std)\end{tabular} & \begin{tabular}[c]{@{}c@{}}CF\textsubscript{MSE}\\ Avg (Std)\end{tabular} \\
              & 0.8s                     & 1.6s                    & 1.6s                                                         & 1.6s                                 \\ \midrule
BA-PTP\textsubscript{BB+BO+P} - C & 243$\pm$7 & 792$\pm$22         & 692$\pm$18                                                   & 2025$\pm$66                          \\
\textbf{BA-PTP\textsubscript{BB+BO+P} - I} & \textbf{233$\pm$6} & \textbf{768$\pm$32} & \textbf{680$\pm$35}                       & \textbf{1966$\pm$91}                 \\ \midrule
BA-PTP\textsubscript{BB+BO+HO+P} - C & 246$\pm$9 & 807$\pm$16      & 704$\pm$15                                                   & 2086$\pm$37                          \\
\textbf{BA-PTP\textsubscript{BB+BO+HO+P} - I} & \textbf{236$\pm$2} & \textbf{777$\pm$13} & \textbf{685$\pm$14}                    & \textbf{2003$\pm$46}                 \\ \bottomrule
\end{tabular}
\end{center}
\end{table}

\paragraph{Results on the PIE dataset}
In this section, we evaluate our proposed method on the PIE dataset \cite{rasouli2019:pie} using inferred values for body orientation and pose of pedestrians as described in section~\ref{sec_implementation}.
The results for PIE\textsubscript{traj} \cite{rasouli2019:pie} and SGNet \cite{wang2022:sgnet} are taken from the respective publications.
As there are no annotations for the ego-vehicle's yaw rate for the PIE dataset, we cannot perform the exact same experiments as on ECP-Intention.
However, the authors of PIE instead provide the ego vehicle's yaw, which we also use in an experiment (BA-PTP\textsubscript{BB+Yaw}).
The results are presented in Table~\ref{table_pie_results}.

When comparing the results of PIE\textsubscript{traj+speed} with our model that additionally uses yaw, we do not see the significant improvements we did on ECP-Intention when using the yaw rate.
Furthermore, our model that solely uses bounding box and speed performs better across all metrics.
These results imply that the incorporation of the ego-vehicle's yaw is not important on PIE.
This could be explained by the fact that the scenes in that dataset mostly cover scenarios where the vehicle drives in a straight line or stands still, in contrast to ECP-Intention.
As a consequence, we do not use the yaw for the behavior-aware variants of our method, as we found this to result in worse performance.

Table~\ref{table_pie_results} shows that using inferred behavioral features in addition to bounding box and ego-vehicle speed improves the prediction of pedestrians' future trajectories also on the PIE dataset.
Adding body orientation increases the prediction performance by a small margin (3\% in MSE-1.5s) while using pose information improves performance a lot (12\% in MSE-1.5s).
The best performing model uses both body orientation and pose in addition to bounding boxes and ego-vehicle speed.
Compared to BA-PTP\textsubscript{BB+P} we improve across all metrics by a few points.
%
%
Compared to the baseline PIE\textsubscript{traj+speed} we improve by 15\% in terms of MSE-1.5s.
Our model achieves the best results for MSE-1.5s, C\textsubscript{MSE} and CF\textsubscript{MSE} outperforming SGNet by 4\%, 7\% and 14\%, respectively.

\begin{table}[tb]
\caption{Trajectory prediction results on the PIE dataset
\\ (BB=Bounding Boxes, BO=Body Orientation, P=Pose)}\label{table_pie_results}
\begin{center}
\begin{tabular}{@{}lccccc@{}}
\toprule
\multirow{3}{*}{Model} & \multicolumn{3}{c}{\begin{tabular}[c]{@{}c@{}}MSE\\ Avg (Std)\end{tabular}} & \begin{tabular}[c]{@{}c@{}}C\textsubscript{MSE}\\ Avg (Std)\end{tabular} & \begin{tabular}[c]{@{}c@{}}CF\textsubscript{MSE}\\ Avg (Std)\end{tabular} \\
            & 0.5s                     & 1s                      & 1.5s                    & 1.5s                                & 1.5s                    \\ \midrule
PIE\textsubscript{traj} \cite{rasouli2019:pie} & 58 & 200        & 636                     & 596                                 & 2477                    \\
PIE\textsubscript{traj+speed} \cite{rasouli2019:pie} & 60$\pm$2 & 173$\pm$4 & 498$\pm$9    & 450$\pm$8                           & 1782$\pm$40             \\
SGNet \cite{wang2022:sgnet} & \textbf{34} & \textbf{133}         & 442                     & 413                                 & 1761                    \\ \midrule
BA-PTP\textsubscript{BB+Yaw} & 58$\pm$4 & 175$\pm$8              & 508$\pm$10              & 471$\pm$10                          & 1876$\pm$29             \\ \midrule
BA-PTP\textsubscript{BB} & 55$\pm$1    & 166$\pm$2               & 487$\pm$7               & 451$\pm$6                           & 1814$\pm$32             \\
BA-PTP\textsubscript{BB+BO} & 59$\pm$1 & 167$\pm$2               & 470$\pm$6               & 434$\pm$6                           & 1723$\pm$28             \\
BA-PTP\textsubscript{BB+P} & 56$\pm$1  & 153$\pm$1               & 426$\pm$4               & 390$\pm$4                           & 1527$\pm$22             \\
BA-PTP\textsubscript{BB+BO+P} & 56$\pm$2 & 151$\pm$3             & \textbf{420$\pm$10}     & \textbf{383$\pm$10}                 & \textbf{1513$\pm$47}    \\ \bottomrule
\end{tabular}
\end{center}
\end{table}

\begin{figure*}[tb]
\centering
\begin{tabular}{ccc}
\includegraphics[width=0.31\textwidth]{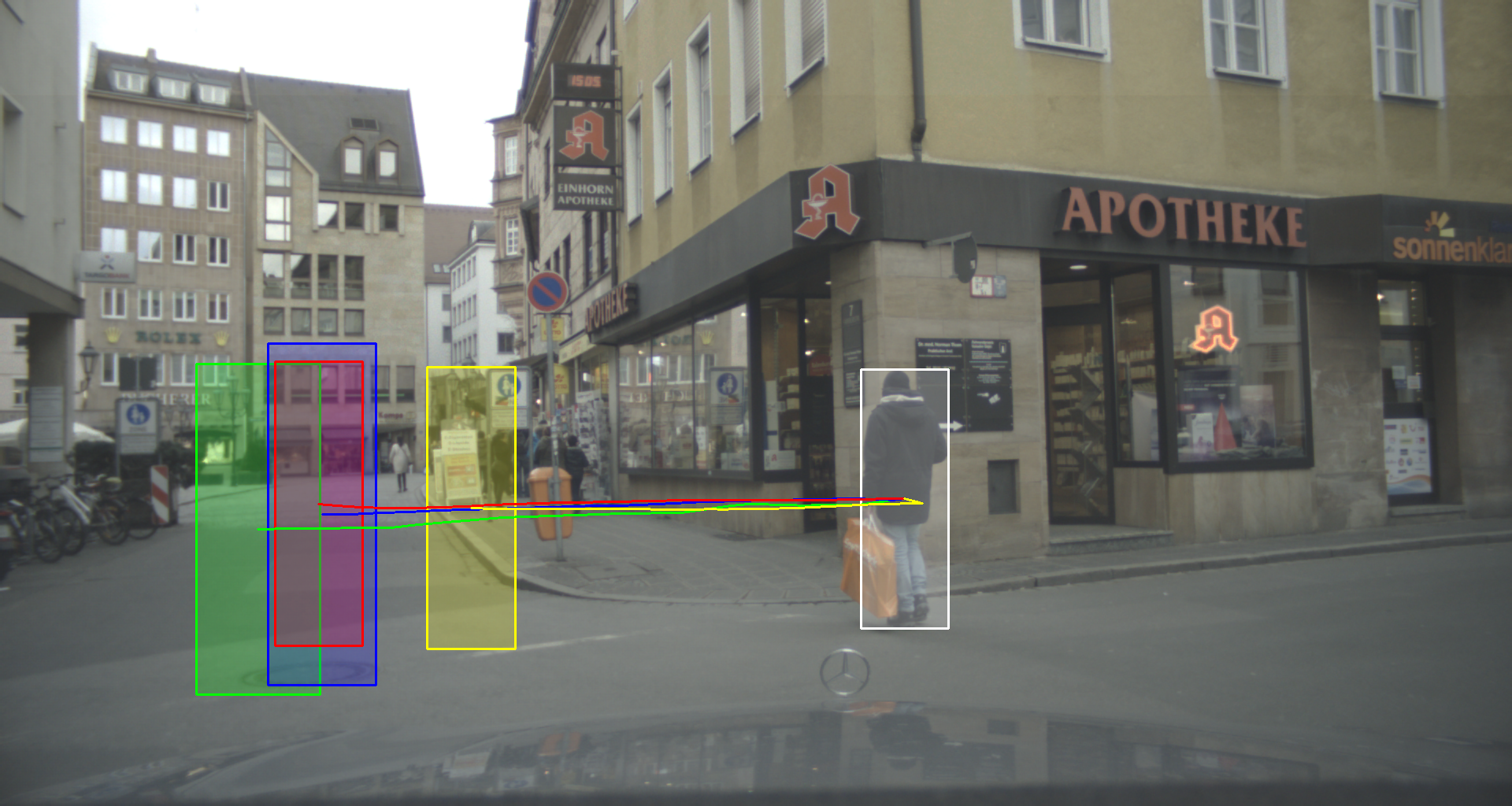} & \includegraphics[width=0.31\textwidth]{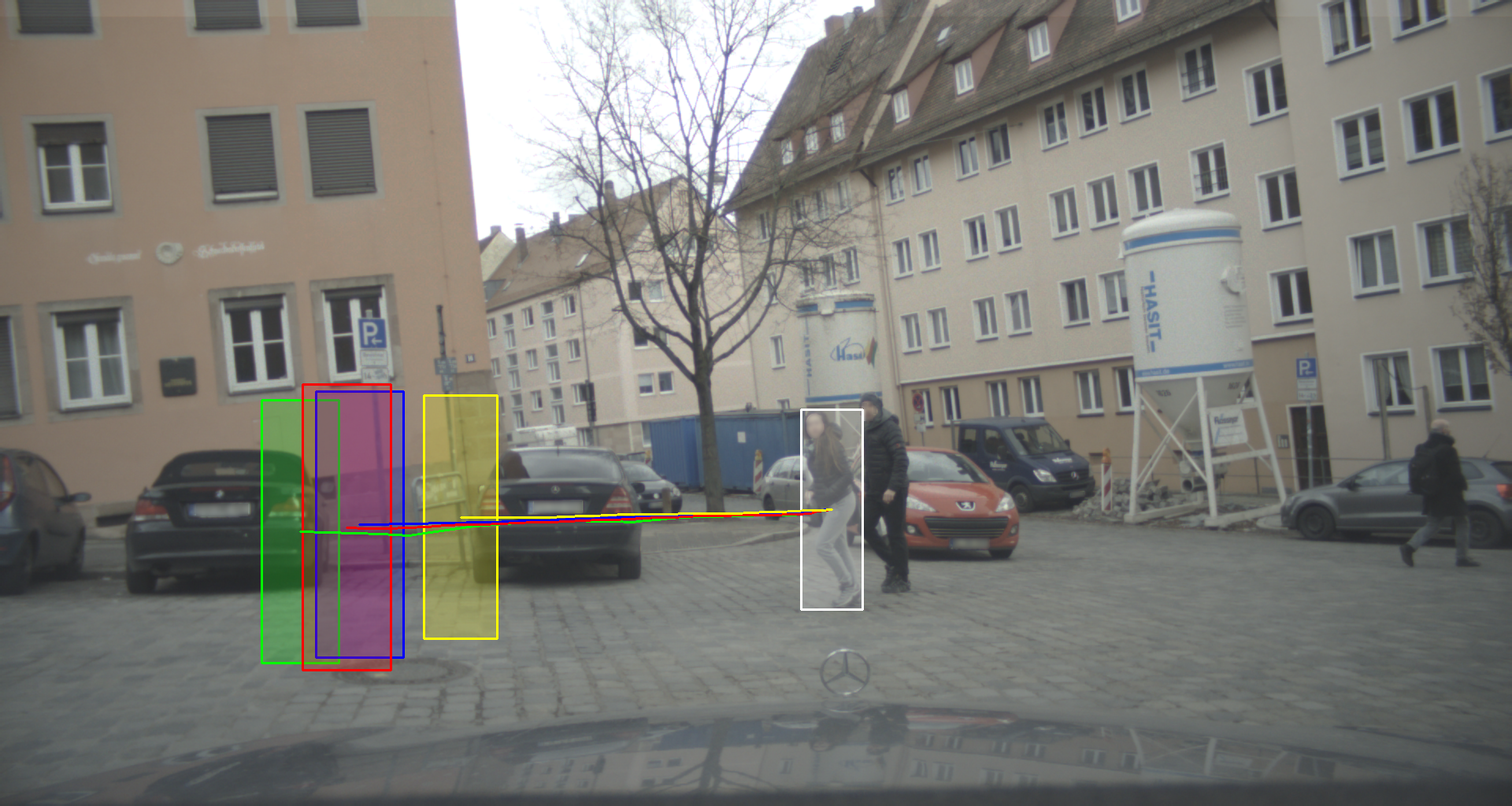}  & \includegraphics[width=0.31\textwidth]{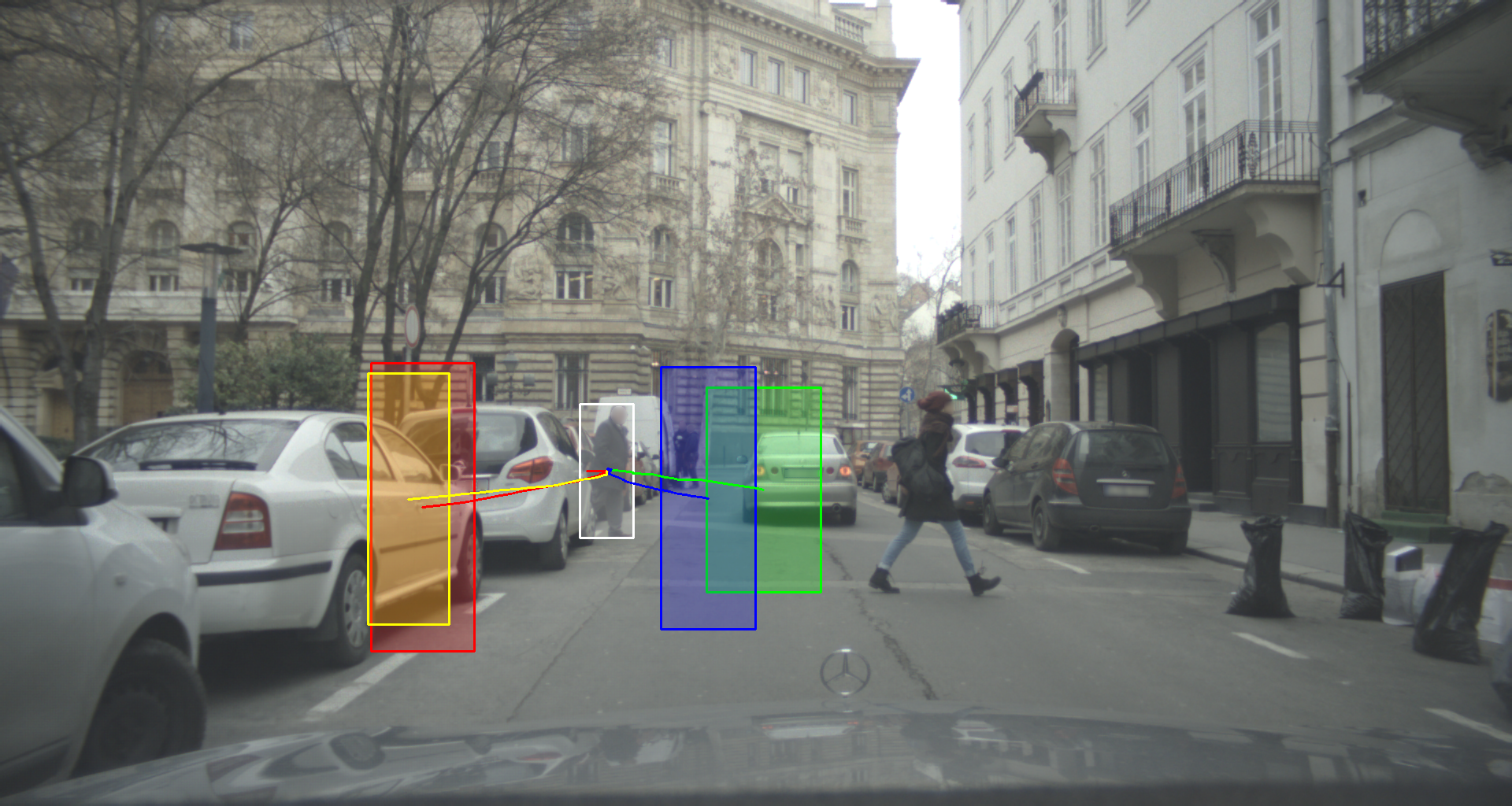} \\
(a) & (b) & (c) \\
\includegraphics[width=0.31\textwidth]{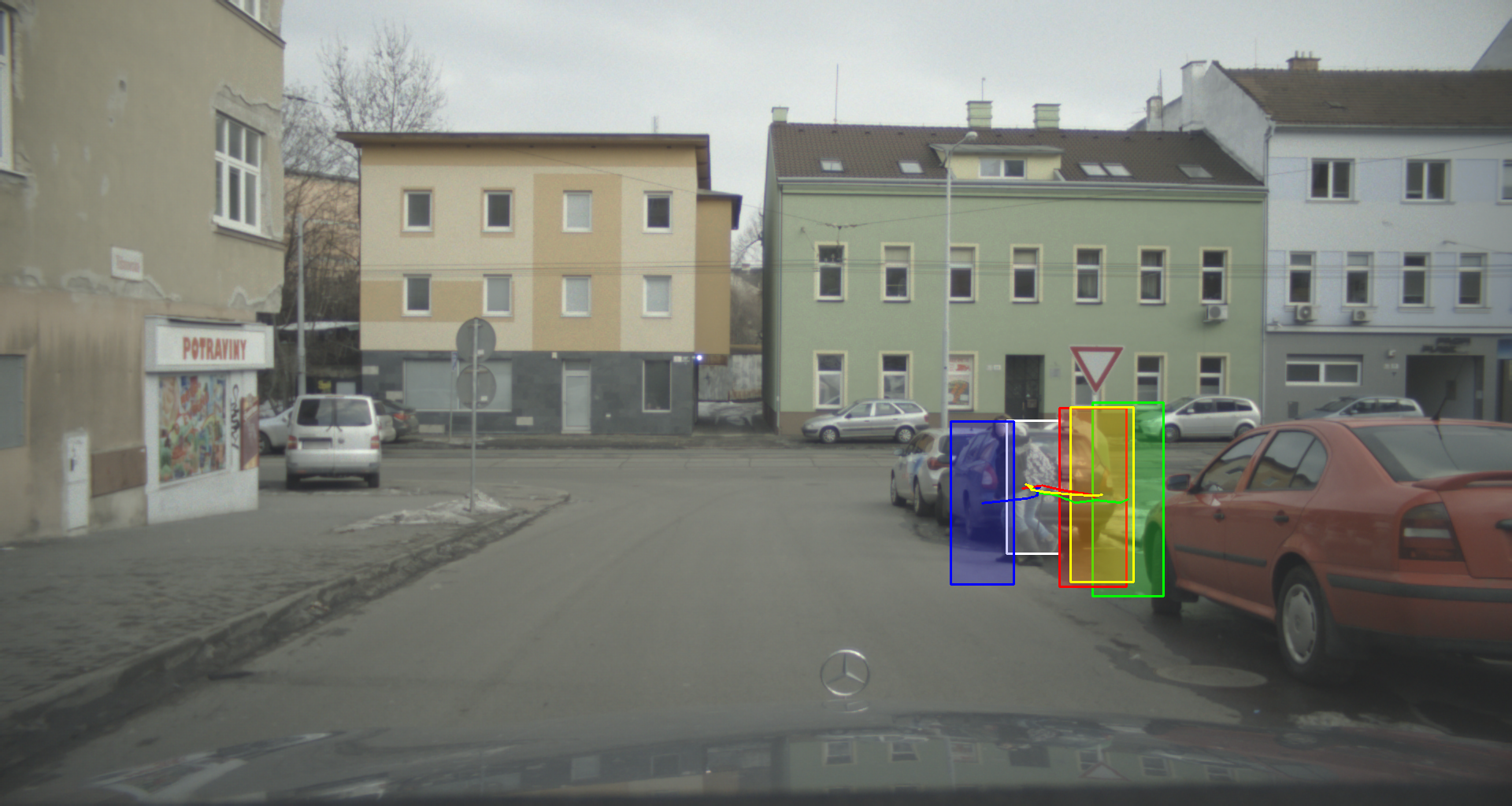} & \includegraphics[width=0.31\textwidth, trim= 0 56 0 0, clip=true]{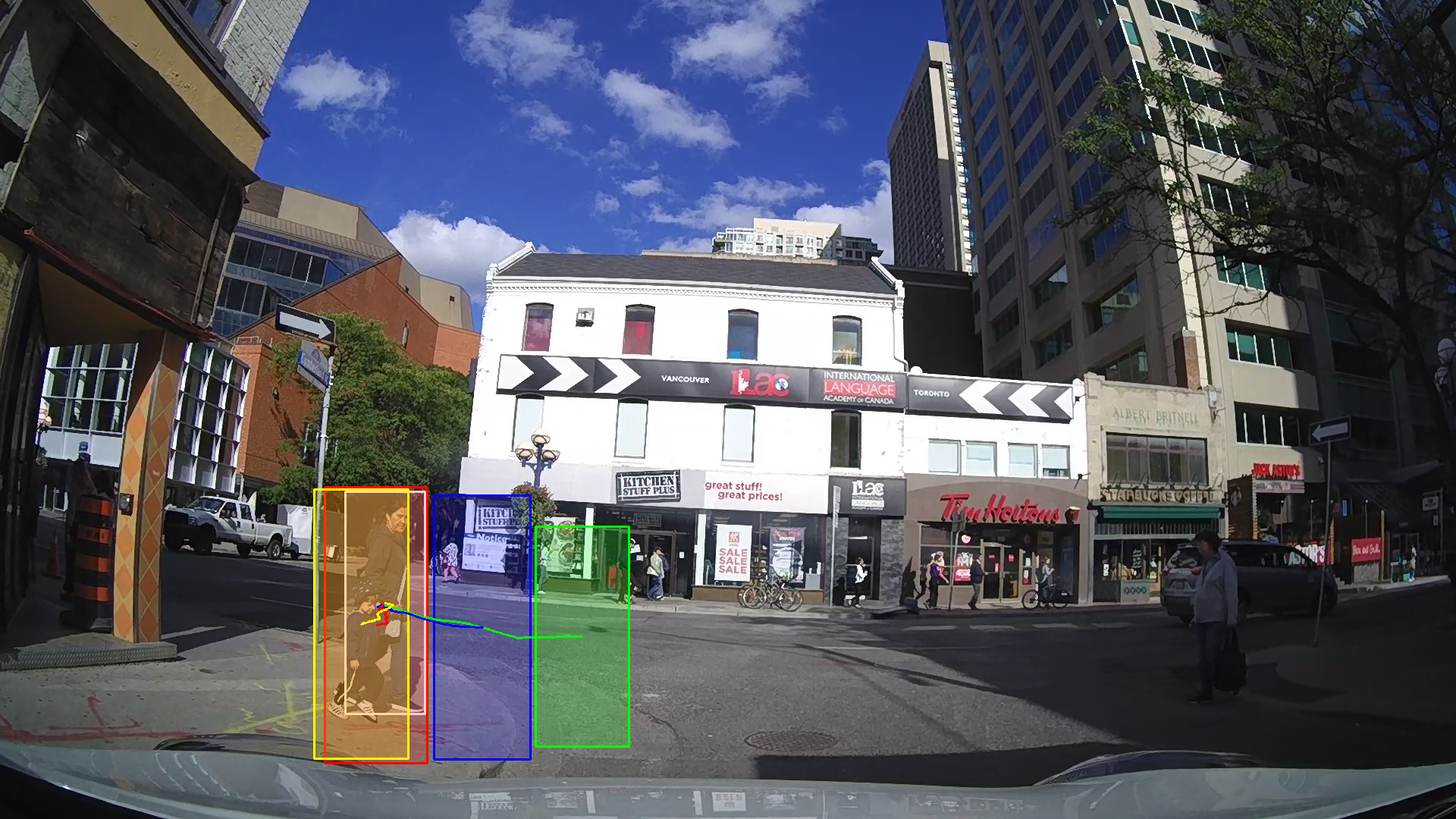} & \includegraphics[width=0.31\textwidth, trim= 0 56 0 0, clip=true]{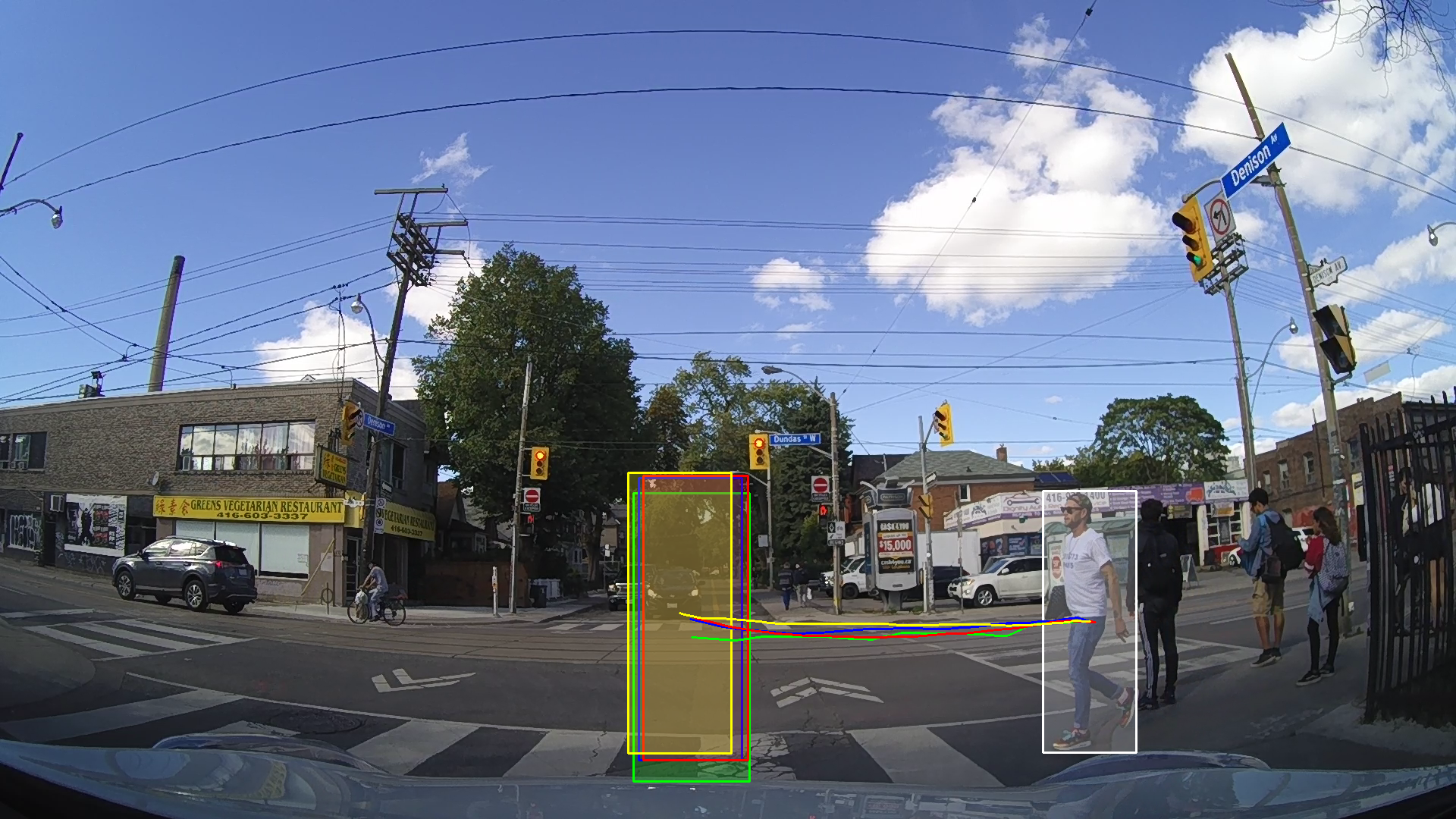} \\
(d) & (e) & (f)
\end{tabular}
\caption{Qualitative results on the ECP-Intention dataset (images a - d) and on the PIE dataset (images e and f).
The images show the last observed timestep with the last observed bounding box (white box) as well as predictions into the future (1.6s for ECP-Intention and 1.5s for PIE).
Visualized are the final predicted bounding box and a line connecting the centers of the predicted bounding boxes for the intermediate timesteps.
Ground truth is displayed in \textcolor{green}{green}.
Predictions of PIE\textsubscript{traj+speed} in \textcolor{yellow}{yellow}, our behavior-aware model BA-PTP\textsubscript{BB+BO+P} in \textcolor{blue}{blue} and our bounding box only model BA-PTP\textsubscript{BB} in \textcolor{red}{red}.
}
\label{fig_qualitative_results}
\end{figure*}

\paragraph{Qualitative results}
Fig.~\ref{fig_qualitative_results} shows qualitative results of our proposed models on ECP-Intention as well as on PIE.
We compare our behavior-aware model BA-PTP\textsubscript{BB+BO+P} with two other models:
An ablation of our model solely relying on bounding boxes and odometry information BA-PTP\textsubscript{BB} and with PIE\textsubscript{traj+speed}.

Images (a) and (b) cover two example scenarios where the ego-vehicle is making a turn.
In such cases, both of our models are able to compensate for the ego-motion of the camera due to the incorporation of the ego-vehicle's yaw rate, whereas PIE\textsubscript{traj+speed} lacks behind.
Images (c) and (e) show examples where BA-PTP\textsubscript{BB+BO+P} correctly forecasts a motion change, i.e. that the pedestrians in both images will walk on the street.
The other two models fail to predict the correct future trajectory.
This can be explained by the pedestrians' orientations towards the street and shows that behavioral features such as body orientation as well as pose can help to improve pedestrian trajectory prediction.

We also show a failure case in image (d) on the ECP-Intention dataset, where the pedestrian suddenly stops crossing the street.
The models that only use bounding boxes correctly predict that the pedestrian will stand still, whereas BA-PTP\textsubscript{BB+BO+P} predicts that the pedestrian's future location will be on the street.
This failure might be caused within the modality attention module by focusing insufficiently on the bounding box information.
In image (f) we see a pedestrian walking over a zebra crossing with a periodic motion pattern.
All models predict accurate future bounding boxes, since the positional information from past bounding boxes, which all models use, is sufficient here.

%% file: 05_conclusion.tex
\section{Conclusion}
In this paper, we presented a novel approach to pedestrian trajectory prediction from the on-board perspective.
Our method BA-PTP uses behavioral features such as the body and head orientation of pedestrians as well as their pose in addition to bounding boxes.
For each input modality, we use an individual encoding stream and fuse the different outputs to better model the motion history of pedestrians.
This way we explicitly utilize hints that pedestrians give about where they are going to move, which can be inferred from visual observation.
By evaluating our proposed method on two datasets for predicting pedestrian behavior we showed that behavioral features benefit pedestrian trajectory prediction and verified the effectiveness of our independent encoding strategy.
Additionally, we investigated which behavioral features contribute most to the prediction performance based on an ablation study.
From that, we found that the best prediction performance is achieved when using body orientation and pose of pedestrians as inputs in addition to bounding boxes and odometry information.
We have shown a slight decrease in performance when also introducing head orientation.
In future work, we seek to investigate the influence of inferred odometry information from an ego-motion model on the trajectory prediction performance of our models.

%% file: 06_acknowledgment.tex
\section*{Acknowledgment}
This work is a result of the research project @CITY – Automated Cars and Intelligent Traffic in the City.
The project is supported by the Federal Ministry for Economic Affairs and Climate Action (BMWK), based on a decision taken by the German Bundestag. 
The authors are solely responsible for the content of this publication.

%% file: ms.bbl
\begin{thebibliography}{10}
\providecommand{\url}[1]{#1}
\csname url@rmstyle\endcsname
\providecommand{\newblock}{\relax}
\providecommand{\bibinfo}[2]{#2}
\providecommand\BIBentrySTDinterwordspacing{\spaceskip=0pt\relax}
\providecommand\BIBentryALTinterwordstretchfactor{4}
\providecommand\BIBentryALTinterwordspacing{\spaceskip=\fontdimen2\font plus
\BIBentryALTinterwordstretchfactor\fontdimen3\font minus
  \fontdimen4\font\relax}
\providecommand\BIBforeignlanguage[2]{{%
\expandafter\ifx\csname l@#1\endcsname\relax
\typeout{** WARNING: IEEEtran.bst: No hyphenation pattern has been}%
\typeout{** loaded for the language `#1'. Using the pattern for}%
\typeout{** the default language instead.}%
\else
\language=\csname l@#1\endcsname
\fi
#2}}

\bibitem{rudenko2020:survey}
A.~Rudenko, L.~Palmieri, M.~Herman, K.~M. Kitani, D.~M. Gavrila, \emph{et~al.},
  ``Human motion trajectory prediction: A survey,'' \emph{The International
  Journal of Robotics Research}, vol.~39, no.~8, pp. 895--935, 2020.

\bibitem{rasouli2019:pie}
A.~Rasouli, I.~Kotseruba, T.~Kunic, and J.~K. Tsotsos, ``Pie: A large-scale
  dataset and models for pedestrian intention estimation and trajectory
  prediction,'' in \emph{\iccv}, 2019.

\bibitem{bhattacharyya2018:bayesian_lstm}
A.~Bhattacharyya, M.~Fritz, and B.~Schiele, ``Long-term on-board prediction of
  people in traffic scenes under uncertainty,'' in \emph{\cvpr}, 2018, pp.
  4194--4202.

\bibitem{wang2022:sgnet}
C.~Wang, Y.~Wang, M.~Xu, and D.~J. Crandall, ``Stepwise goal-driven networks
  for trajectory prediction,'' \emph{IEEE Robotics and Automation Letters},
  vol.~7, no.~2, pp. 2716--2723, 2022.

\bibitem{yao2021:bitrap}
Y.~Yao, E.~Atkins, M.~Johnson-Roberson, R.~Vasudevan, and X.~Du, ``Bitrap:
  Bi-directional pedestrian trajectory prediction with multi-modal goal
  estimation,'' \emph{IEEE Robotics and Automation Letters}, vol.~6, no.~2, pp.
  1463--1470, 2021.

\bibitem{cao2020:ht_stgcn}
D.~Cao and Y.~Fu, ``Using graph convolutional networks skeleton-based
  pedestrian intention estimation models for trajectory prediction,'' in
  \emph{Journal of Physics: Conference Series}, vol. 1621, no.~1, 2020, p.
  012047.

\bibitem{sui2021:joint_intention_and_traj_based_on_tranformer}
Z.~Sui, Y.~Zhou, X.~Zhao, A.~Chen, and Y.~Ni, ``Joint intention and trajectory
  prediction based on transformer,'' in \emph{\iros}, 2021, pp. 7082--7088.

\bibitem{lorenzo2020:rnn_based_crossing_prediction_using_features}
J.~Lorenzo, I.~Parra, F.~Wirth, C.~Stiller, D.~F. Llorca, \emph{et~al.},
  ``Rnn-based pedestrian crossing prediction using activity and pose-related
  features,'' in \emph{\iv}, 2020, pp. 1801--1806.

\bibitem{yau2021:graph_sim}
T.~Yau, S.~Malekmohammadi, A.~Rasouli, P.~Lakner, M.~Rohani, \emph{et~al.},
  ``Graph-sim: A graph-based spatiotemporal interaction modelling for
  pedestrian action prediction,'' in \emph{\icra}, 2021, pp. 8580--8586.

\bibitem{rasouli2021:biped}
A.~Rasouli, M.~Rohani, and J.~Luo, ``Bifold and semantic reasoning for
  pedestrian behavior prediction,'' in \emph{\iccv}, 2021, pp.
  15\,600--15\,610.

\bibitem{kotseruba2021:pcpa}
I.~Kotseruba, A.~Rasouli, and J.~K. Tsotsos, ``Benchmark for evaluating
  pedestrian action prediction,'' in \emph{\wacv}, 2021, pp. 1258--1268.

\bibitem{yao2021:coupling_intent_and_action}
Y.~Yao, E.~Atkins, M.~Johnson-Roberson, R.~Vasudevan, and X.~Du, ``Coupling
  intent and action for pedestrian crossing behavior prediction,'' in
  \emph{\ijcai}, 2021, pp. 1238--1244.

\bibitem{rasouli2017:jaad_baseline}
A.~Rasouli, I.~Kotseruba, and J.~K. Tsotsos, ``Are they going to cross? a
  benchmark dataset and baseline for pedestrian crosswalk behavior,'' in
  \emph{\iccvw}, 2017, pp. 206--213.

\bibitem{fang2018:pedestrian}
Z.~Fang and A.~M. L{\'o}pez, ``Is the pedestrian going to cross? answering by
  2d pose estimation,'' in \emph{\iv}, 2018, pp. 1271--1276.

\bibitem{cadena2022:pedestrian_graph_plus}
P.~R.~G. Cadena, Y.~Qian, C.~Wang, and M.~Yang, ``Pedestrian graph +: A fast
  pedestrian crossing prediction model based on graph convolutional networks,''
  \emph{\its}, pp. 1--12, 2022, in press.

\bibitem{liu2020:stip}
B.~Liu, E.~Adeli, Z.~Cao, K.-H. Lee, A.~Shenoi, \emph{et~al.}, ``Spatiotemporal
  relationship reasoning for pedestrian intent prediction,'' \emph{IEEE
  Robotics and Automation Letters}, vol.~5, no.~2, pp. 3485--3492, 2020.

\bibitem{varytimidis2018:action}
D.~Varytimidis, F.~Alonso-Fernandez, B.~Duran, and C.~Englund, ``Action and
  intention recognition of pedestrians in urban traffic,'' in \emph{\sitis},
  2018, pp. 676--682.

\bibitem{yagi2018:future}
T.~Yagi, K.~Mangalam, R.~Yonetani, and Y.~Sato, ``Future person localization in
  first-person videos,'' in \emph{\cvpr}, 2018, pp. 7593--7602.

\bibitem{yang2022:predicting_crossing_with_feat_fusion}
D.~Yang, H.~Zhang, E.~Yurtsever, K.~Redmill, and U.~Ozguner, ``Predicting
  pedestrian crossing intention with feature fusion and spatio-temporal
  attention,'' \emph{IEEE Transactions on Intelligent Vehicles}, pp. 1--1,
  2022.

\bibitem{malla2020:titan}
S.~Malla, B.~Dariush, and C.~Choi, ``Titan: Future forecast using action
  priors,'' in \emph{\cvpr}, 2020, pp. 11\,186--11\,196.

\bibitem{herman2021:bosch_paper}
M.~Herman, J.~Wagner, V.~Prabhakaran, N.~Möser, H.~Ziesche, \emph{et~al.},
  ``Pedestrian behavior prediction for automated driving: Requirements,
  metrics, and relevant features,'' \emph{\its}, pp. 1--16, 2021, in press.

\bibitem{kooij2019:context_based_path_prediction}
J.~F. Kooij, F.~Flohr, E.~A. Pool, and D.~M. Gavrila, ``Context-based path
  prediction for targets with switching dynamics,'' \emph{International Journal
  of Computer Vision}, vol. 127, no.~3, pp. 239--262, 2019.

\bibitem{ridel2019:understanding}
D.~A. Ridel, N.~Deo, D.~Wolf, and M.~Trivedi, ``Understanding
  pedestrian-vehicle interactions with vehicle mounted vision: An lstm model
  and empirical analysis,'' in \emph{\iv}, 2019, pp. 913--918.

\bibitem{alahi2016:social_lstm}
A.~Alahi, K.~Goel, V.~Ramanathan, A.~Robicquet, L.~Fei-Fei, \emph{et~al.},
  ``Social lstm: Human trajectory prediction in crowded spaces,'' in
  \emph{\cvpr}, 2016, pp. 961--971.

\bibitem{sadeghian2019:sophie}
A.~Sadeghian, V.~Kosaraju, A.~Sadeghian, N.~Hirose, H.~Rezatofighi,
  \emph{et~al.}, ``Sophie: An attentive gan for predicting paths compliant to
  social and physical constraints,'' in \emph{\cvpr}, 2019, pp. 1349--1358.

\bibitem{lee2017:desire}
N.~Lee, W.~Choi, P.~Vernaza, C.~B. Choy, P.~H. Torr, \emph{et~al.}, ``Desire:
  Distant future prediction in dynamic scenes with interacting agents,'' in
  \emph{\cvpr}, 2017, pp. 336--345.

\bibitem{salzmann2020:trajectron++}
T.~Salzmann, B.~Ivanovic, P.~Chakravarty, and M.~Pavone, ``Trajectron++:
  Dynamically-feasible trajectory forecasting with heterogeneous data,'' in
  \emph{\eccv}, 2020, pp. 683--700.

\bibitem{zhao2021:where_are_you_heading}
H.~Zhao and R.~P. Wildes, ``Where are you heading? dynamic trajectory
  prediction with expert goal examples,'' in \emph{\iccv}, 2021, pp.
  7629--7638.

\bibitem{neumann2021:pedestrian_and_ego_veh_traj}
L.~Neumann and A.~Vedaldi, ``Pedestrian and ego-vehicle trajectory prediction
  from monocular camera,'' in \emph{\cvpr}, 2021, pp. 10\,204--10\,212.

\bibitem{yin2021:multimodal_transformer}
Z.~Yin, R.~Liu, Z.~Xiong, and Z.~Yuan, ``Multimodal transformer networks for
  pedestrian trajectory prediction.'' in \emph{\ijcai}, 2021, pp. 1259--1265.

\bibitem{braun2019:ecp}
M.~Braun, S.~Krebs, F.~Flohr, and D.~M. Gavrila, ``{EuroCity Persons: A} novel
  benchmark for person detection in traffic scenes,'' \emph{\pami}, vol.~41,
  no.~8, pp. 1844--1861, 2019.

\bibitem{kingma2014:adam}
D.~P. Kingma and J.~Ba, ``Adam: A method for stochastic optimization,''
  \emph{arXiv preprint arXiv:1412.6980}, 2014.

\bibitem{kumar2021:pose_ssd}
C.~Kumar, J.~Ramesh, B.~Chakraborty, R.~Raman, C.~Weinrich, \emph{et~al.},
  ``Vru pose-ssd: Multiperson pose estimation for automated driving,'' in
  \emph{Proceedings of the AAAI Conference on Artificial Intelligence},
  vol.~35, no.~17, 2021, pp. 15\,331--15\,338.

\bibitem{braun2016:pose_rcnn}
M.~Braun, Q.~Rao, Y.~Wang, and F.~Flohr, ``Pose-rcnn: Joint object detection
  and pose estimation using 3d object proposals,'' in \emph{\itsc}, 2016, pp.
  1546--1551.

\bibitem{braun2021:simple_pair_pose}
M.~Braun, F.~B. Flohr, S.~Krebs, U.~Kre{\ss}el, and D.~M. Gavrila, ``Simple
  pair pose-pairwise human pose estimation in dense urban traffic scenes,'' in
  \emph{\iv}, 2021, pp. 1545--1552.

\bibitem{wang2021:urbanpose}
S.~Wang, D.~Yang, B.~Wang, Z.~Guo, R.~Verma, \emph{et~al.}, ``Urbanpose: A new
  benchmark for vru pose estimation in urban traffic scenes,'' in \emph{\iv},
  2021, pp. 1537--1544.

\end{thebibliography}
